\pdfoutput=1

\documentclass[11pt]{article}

\usepackage{acl}

\usepackage{times}
\usepackage{url}
\usepackage{latexsym}
\usepackage{natbib}
\usepackage{layout}
\usepackage[hang,marginal]{footmisc}

\usepackage{inconsolata}
\usepackage{multirow}
\usepackage{multicol}
\usepackage{booktabs}
\usepackage{hyperref}
\usepackage{subcaption}
\usepackage{graphicx}

\usepackage{amsmath}
\usepackage{amssymb}
\usepackage{mathtools}
\usepackage{amsthm}
\usepackage{caption}
\usepackage[capitalize,noabbrev]{cleveref}
\usepackage{amsfonts}

\usepackage[T1]{fontenc}

\usepackage[utf8]{inputenc}

\usepackage{microtype}

\title{\bf DePA: Improving Non-autoregressive Machine Translation with Dependency-Aware Decoder}

\author{
Jiaao Zhan$^{1}$, Qian Chen, Boxing Chen, Wen Wang, Yu Bai$^{1}$, Yang Gao$^{1}$\thanks{~~Corresponding Author}  \\
    $^{1}$School of Computer Science and Technology,\\
    Beijing Institute of Technology, Beijing, China \\
  \texttt{jiaao\_zhan@163.com}\\ 
  \texttt{\{lukechan1231,chenboxing,wwang.969803\}@gmail.com} \\
  \texttt{\{yubai,gyang\}@bit.edu.cn} \\
 }

\begin{document}
\maketitle
\begin{abstract}
Non-autoregressive machine translation (NAT) models have lower translation quality than autoregressive translation (AT) models because NAT decoders do not depend on previous target tokens in the decoder input. We propose a novel and general \textbf{Dependency-Aware Decoder (DePA)} to enhance target dependency modeling in the decoder of fully NAT models from two perspectives: decoder self-attention and decoder input. First, we propose an autoregressive forward-backward pre-training phase before NAT training, which enables the NAT decoder to gradually learn bidirectional target dependencies for the final NAT training. Second, we transform the decoder input from the source language representation space to the target language representation space through a novel attentive transformation process, which enables the decoder to better capture target dependencies. DePA can be applied to any fully NAT models. Extensive experiments show that DePA consistently improves highly competitive and state-of-the-art fully NAT models on widely used WMT and IWSLT benchmarks by up to \textbf{1.88} BLEU gain, while maintaining the inference latency comparable to other fully NAT models.\footnote{We released our code at: \url{https://github.com/zhanjiaao/NAT\_DePA}.}

\end{abstract}
\section{Introduction}
\label{sec:intro}
Autoregressive translation (AT) systems achieve state-of-the-art (SOTA) performance for neural machine translation (NMT) and Transformer~\citep{vaswani2017attention} encoder-decoder is the prevalent architecture. In AT systems, each generation step depends on previously generated tokens, resulting in high inference latency when output is long. Non-autoregressive translation (NAT) models~\citep{gu2018non} significantly accelerate inference by generating all target tokens independently and simultaneously. However, this independence assumption leads to degradation in accuracy compared to AT models, as NAT models cannot properly learn target dependencies. \textit{Dependency} in prior works and our work takes its standard definition in NLP, i.e., syntactic relations between words in a sentence.

The mainstream NAT models fall into two categories: iterative NAT models and fully NAT models.
Iterative NAT models~\citep{gu2019levenshtein,ghazvininejad2019mask,lee2020deterministic} improve translation accuracy by iteratively refining translations at the expense of slower decoding speed. 
In contrast, fully NAT models~\citep{gu2018non,DBLP:conf/acl/BaoZHWQDCL22} have great latency advantage over AT models 
by making parallel predictions with a single decoding round, but they suffer from lower translation accuracy. In this paper, we aim at improving the translation accuracy of \textbf{fully NAT models} while preserving their latency advantage.

\begin{table*}[t]
\small
\centering
\resizebox{1.0\linewidth}{!}{
\begin{tabular}{c|c|c|c}
\toprule
& \multirow{1}*{\textbf{Under-Translation}}&\multirow{1}*{\textbf{Over-Translation}}&\multirow{1}*{\textbf{Wrong Lexical Choice}}\\
\midrule
\multirow{1}*{Source} & \multirow{1}*{Wir haben es \underline{\textbf{weltweit}} in 300 Gemeinden gemacht.}&\multirow{1}*{Einige leute wollten ihn einfach König nennen} & \multirow{1}*{Woher komme ich ? Wer bin ich ?}\\
\midrule
\multirow{1}*{Target Reference} &\multirow{1}*{We 've done it in 300 communities around the world.} & \multirow{1}*{Some people just wanted to call him King .}& \multirow{1}*{Where am I from ? Who am I ?}\\
\midrule
\multirow{1}*{\bf{F-NAT}} & \multirow{1}*{We did it the world in 300 communities.}
& \multirow{1}*{Some people just wanted to call {\color{red}him him} king.}
&\multirow{1}*{{\color{blue}How} do I come from ? Who am I ?} \\
\midrule
\multirow{1}*{\bf{FB-NAT}} & \multirow{1}*{We 've done it in 300 communities around the world.}& \multirow{1}*{Some people just wanted to call him king.} & \multirow{1}*{Where do I come from? Who am I ?}
\\
\bottomrule
\end{tabular}}
\caption{\small{Case studies of our proposed \textbf{FBD} approach on the highly competitive fully NAT model GLAT~\citep{qian2020glancing} for alleviating three types of multi-modality errors on the IWSLT16 DE-EN validation set. Repetitive tokens are in red. Source words that are not semantically translated are in bold and underlined. Wrong lexical choices (for polysemous words) and redundant words are in blue. 
\textbf{F-NAT} denotes only modeling forward dependencies while \textbf{FB-NAT} denotes modeling both forward and backward dependencies, the same as the models in Table~\ref{tab:combine}. Case studies of our proposed \textbf{IT} approach are in Appendix.}}
\label{tab:case_study_fbd}
\end{table*}

Previous research~\citep{gu2020fully} argues that reducing dependencies is crucial for training a fully NAT model effectively, as it allows the model to more easily capture target dependencies. However, dependency reduction limits the performance upper bound of fully NAT models, since models may struggle to generate complex sentences. Previous studies show that multi-modality~\citep{ran2020learning} is the main problem that NAT models suffer from~\citep{huang2021non,DBLP:conf/acl/BaoZHWQDCL22}, i.e., the target tokens may be generated based on different possible translations, often causing over-translation (token repetitions), under-translation (source words not translated), and wrong lexical choice for polysemous words. 
Table~\ref{tab:case_study_fbd} Row3 shows all three multi-modality error types from the highly competitive fully NAT model GLAT~\citep{qian2020glancing} with modeling only forward dependency (F-NAT) in our experiments. We observe that \textbf{lack of complete dependency modeling could cause multi-modality errors}. For example, for the source text (in German) ``\textit{Woher komme ich?}'' in the last column of Table~\ref{tab:case_study_fbd}, ``\textit{Woher}'' means both ``\textit{where}'' and ``\textit{how}''. The NAT model modeling only forward dependency (F-NAT) incorrectly translates ``\textit{woher}'' into ``\textit{how}'' and outputs ``\textit{How do I come from?}''; whereas the model modeling both forward and backward dependency (\textbf{FB-NAT}) translates it correctly into ``\textit{Where do I come from?}''.
Therefore, instead of dependency reduction, we propose a novel and general Dependency-Aware Decoder (\textbf{DePA}), which enhances the learning capacity of fully NAT models and enables them to learn \emph{complete} and \emph{complex} forward and backward target dependencies in order to alleviate the multi-modality issue.

Firstly, we enhance the NAT decoder to learn complete target dependencies by exploring decoder self-attention. We believe that previous works~\citep{guo2020fine} incorporating only forward dependency modeled by AT models into NAT models are inadequate to address multi-modality. Therefore, we propose an effective forward-backward dependency modeling approach, denoted by \textbf{FBD}, as an auto-aggressive forward-backward pre-training phase before NAT training, using curriculum learning. The FBD approach implements \textit{triangular attention masks} and takes different decoder inputs and targets \textit{in a unified framework} to train the model to attend to previous or future tokens and learn both forward or backward dependencies.

Secondly, we enhance target dependency modeling within the NAT decoder from the perspective of the decoder input. 
Most prior NAT models~\citep{gu2018non,wang2019non,wei2019imitation} use a copy of the source text embedding as the decoder input, which is independent from the target representation space and hence makes target dependency modeling difficult. 
We transform the initial decoder input from the source language representation space to the target language representation space through a novel attentive transformation process, denoted by \textbf{IT}. 
Previous works on transforming the decoder input cannot guarantee that the decoder input is in the exact target representation space, resulting in differences from the true target-side distribution. Our proposed IT ensures that the decoder input is in the \textit{exact} target representation space hence 
enables the model to better capture target dependencies.

Our contributions can be summarized as follows: (1) We propose a novel and general \textbf{Dependency-Aware Decoder (DePA)} for fully NAT models. 
For DePA, we propose a novel approach \textbf{FBD} for learning both forward and backward dependencies in NAT decoder, through which the target dependencies can be better modeled. \textbf{To the best of our knowledge, our work is the first to successfully model both forward and backward target-side dependencies \textit{explicitly} for fully NAT models}.
We also propose a novel decoder input transformation approach (\textbf{IT}). IT could ease target-side dependency modeling and enhance the effectiveness of FBD. DePA is \textbf{model-agnostic} and can be applied to any fully NAT models.  (2) Extensive experiments on WMT and IWSLT benchmarks demonstrate that our DePA consistently improves the representative vanilla NAT model~\citep{gu2018non}, the highly competitive fully NAT model GLAT~\citep{qian2020glancing} and the current SOTA of fully NAT models, CTC w/ DSLP \& Mixed Training (denoted by \textbf{CTC-DSLP-MT})~\citep{huang2021non} (\textbf{DSLP} denotes Deep Supervision and Layer-wise Prediction), by up to \textbf{+0.85} BLEU on the SOTA CTC-DSLP-MT, \textbf{+1.88} BLEU on GLAT, and \textbf{+2.2} BLEU on vanilla NAT, while reserving inference latency as other fully NAT models, about $\mathbf{{15}\times}$ speed-up over AT models. Experiments show that DePA
achieves greater BLEU gains with less speed-up loss than DSLP when applied to various fully NAT models.

\section{Related Work}
\paragraph{Forward and Backward Dependencies}
Prior works explore bidirectional decoding to improve modeling of both forward and backward dependencies in phrase-based statistical MT~\citep{DBLP:conf/emnlp/FinchS09} and RNN-based MT~\citep{DBLP:conf/aaai/ZhangSQLJW18}.
For NAT, \citet{guo2020fine} and \citet{wei2019imitation} use forward auto-regressive models to guide NAT training.
\citet{DBLP:conf/ijcai/LiuRTZQZL20} introduces an intermediate semi-autoregressive translation task to smooth the shift from AT training to NAT training. However, backward dependencies are rarely investigated in NAT.

\paragraph{Decoder Input of Fully NAT Models} 
The decoder input of AT models consists of previously generated tokens. However, selecting appropriate decoder input for fully NAT models could be challenging. Most prior NAT models~\citep{gu2018non,wang2019non,wei2019imitation} use uniform copy~\citep{gu2018non} or soft copy~\citep{wei2019imitation} of the source text embedding as the decoder input, which
is independent of the target representation space hence hinders target dependency modeling. Methods such as GLAT~\citep{qian2020glancing} and \citep{guo2020fine, guo2020jointly} attempt to make the NAT decoder input similar to the target representation space
by substituting certain positions in the decoder input with the corresponding target embedding. However, this creates a mismatch between training and inference. \citet{guo2019non} uses phrase-table lookup and linear mapping to make the decoder input closer to the target embedding, but this method still causes difference between the decoder input and the real target-side distribution.

\paragraph{Fully NAT Models}
To address multi-modality for fully NAT models, various approaches are proposed. \citet{gu2018non} uses knowledge distillation (KD)~\citep{kim2016sequence} to reduce dataset complexity. \citet{libovicky2018end} and \citet{saharia2020non} use connectionist temporal classification (CTC)~\citep{DBLP:conf/icml/GravesFGS06} for latent alignment. \citet{sun2019fast} utilizes CRFs to model target positional contexts. \citet{kaiser2018fast}, \citet{ma2019flowseq} and \citet{shu2020latent} incorporate latent variables to guide generation, similar to VAEs~\citep{kingma2013auto}. 
\citet{guo2020incorporating} initializes NAT decoders with pretrained language models.
\citet{huang2021non} proposes CTC with Deep Supervision and Layer-wise Prediction and Mixed Training (\textbf{CTC-DSLP-MT}), setting new SOTA for fully NAT models on WMT benchmarks. DA-Transformer~\citep{DBLP:journals/corr/abs-2205-07459} represents hidden states in a directed acyclic graph to capture dependencies between tokens and generate multiple possible translations. In contrast, our DePA utilizes forward-backward pre-training and a novel attentive transformation of decoder input to enhance target dependency modeling. Under same settings and with KD,  DA-Transformer performs only comparably to CTC-DSLP-MT; however, performance of DA-Transformer benefits notably from \textit{Transformer-big} for KD while CTC-DSLP-MT uses \textit{Transformer-base} for KD. DDRS w/ NMLA~\citep{shao2022non} benefits greatly from using diverse KD references while CTC-DSLP-MT uses only a single KD reference.  Hence, \textbf{CTC-DSLP-MT is still the current SOTA for fully NAT models on WMT benchmarks}.

\paragraph{Non-autoregressive Models}
Besides fully NAT models, iterative NAT models are proposed such as iterative refinement of target sentences~\citep{lee2020deterministic}, masking and repredicting words with low probabilities~\citep{ghazvininejad2019mask}, edit-based methods to iteratively modify decoder output~\citep{stern2019insertion,gu2019levenshtein}, and parallel refinement of every token~\citep{kasai2020non}. Iterative NAT models improve translation accuracy at the cost of slower speed. 
Non-autoregressive models are practically important due to high efficiency. Other than MT, they are applied to various tasks such as image captioning~\citep{gao2019masked}, automatic speech recognition~\citep{chen2019listen}, and text-to-speech synthesis~\citep{oord2018parallel}. 

\section{Methodology}
\subsection{Problem Formulation}
\label{formulation}
NMT can be formulated as a sequence-to-sequence generation problem.
 Given a sequence $X=\{x_1,...,x_N\}$ in the source language, a sequence $Y=\{y_1,...,y_T\}$ in the target language is generated following the conditional probability $P(Y|X)$. 
NAT models are proposed to speed up generation by decoding all the target tokens in parallel, using conditional independent factorization as:
\begin{equation}
\small
\label{eq:NAT_prob}
    P_{NA}(Y|X)=P_L(T|x_{1:N})\cdot\prod^T_{t=1}P(y_t|x_{1:N};\theta)
\end{equation}
\noindent where the target sequence length $T$ is modeled by the conditional distribution $P_L$, and dependence on previous target tokens is removed. 
Compared to AT models, NAT models speed up inference significantly at the expense of translation quality, because the conditional independence assumption in Eq.\ref{eq:NAT_prob} enables parallel processing but lacks explicit modeling of dependency between target tokens. 
 To enhance target dependency modeling, we propose two innovations as  incorporating both forward and backward dependency modeling into the training process (Section~\ref{sec:fbd}) and transforming the decoder input into the target representation space (Section~\ref{sec:IT}).

\begin{figure}[t]
    \centering
    \includegraphics[width=0.45\textwidth]{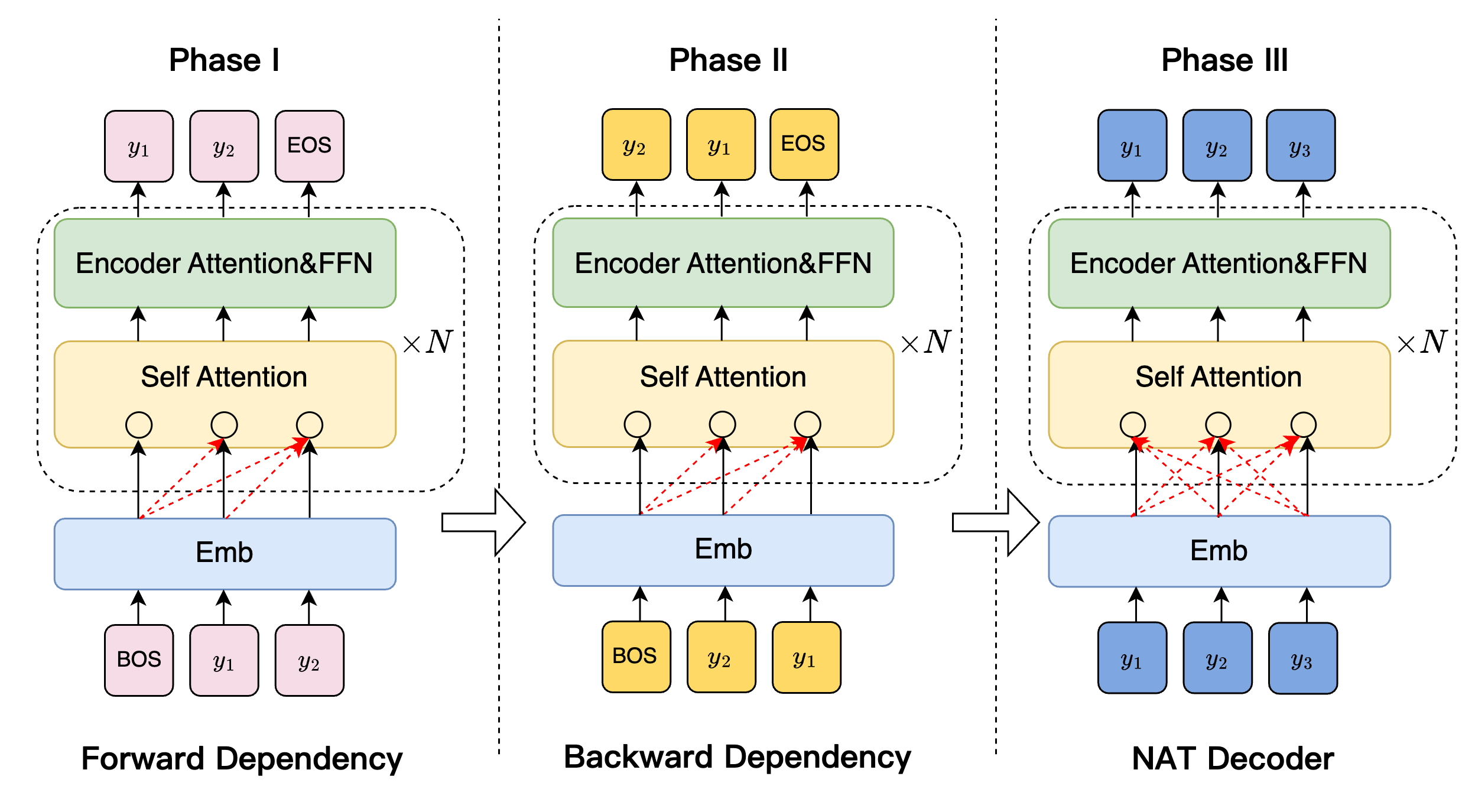}
    \caption{\small{The proposed forward-backward dependency modeling (\textbf{FBD}) with triangular attention masks in a unified framework. The red dashed lines indicate the attention masks.  We use different colors to highlight the difference of inputs and targets in each phase. }}
    \label{fig:fbd}
\end{figure}

\subsection{Target Dependency Modeling with Curriculum Learning (FBD)}
\label{sec:fbd}
Prior work~\citep{guo2020fine} utilizes forward dependency in AT models to initialize model parameters for NAT.
However, as discussed in Section~\ref{sec:intro}, for fully NAT models, only modeling forward dependency is inadequate for addressing the multi-modality problem~\citep{DBLP:conf/emnlp/FinchS09,DBLP:conf/aaai/ZhangSQLJW18} (the Row for F-NAT in Table~\ref{tab:case_study_fbd}).
Our innovations include incorporating both forward and backward dependency modeling into NAT models, via \textit{triangular attention masks} in a \textit{unified framework} through \textit{curriculum learning} (Figure~\ref{fig:fbd}), and investigating efficacy of different curricula. In Figure~\ref{fig:fbd}, the \textbf{NAT decoder} phase denotes standard NAT training of any NAT decoder $Dec$. The \textbf{Forward Dependency} and \textbf{Backward Dependency} phases serve pre-training for NAT training, learning left-to-right and right-to-left dependencies to initialize NAT models with better dependencies. Forward Dependency and Backward Dependency training phases apply \textit{the same upper triangle attention mask} on $Dec$. We use KD data from AT models for each phase but the inputs and the targets are different. The Forward Dependency training phase uses $y_1$ to predict $y_2$ and so on. The Backward Dependency training phase reverses the target sequence and uses $y_2$ to predict $y_1$ and so on. The NAT Training phase uses features of each word to predict the word itself. We make the following hypotheses: (1) Considering the nature of languages, learning forward dependency in Phase 1 is easier for the model for language generation. (2) Modeling backward dependency relies on learned forward dependency knowledge, hence it should be in the second phase. 
In fact, we observe the interesting finding that \textbf{the best curriculum remains forward-backward-forward-NAT (FBF-NAT) for both left-branching and right-branching languages}, proving our hypotheses. We speculate that NAT training may benefit from another forward dependency modeling in Phase 3 because the order of left-to-right is more consistent with characteristics of natural languages, hence adding the second forward dependency modeling after FB (i.e., FBF) smooths the transition to the final NAT training. Detailed discussions are in Section~\ref{ablation}.

\subsection{Decoder Input Transformation (IT) for Target Dependency Modeling}
\label{sec:IT}

\begin{figure*}[thb]
    \centering
    \includegraphics[width=0.80\textwidth]{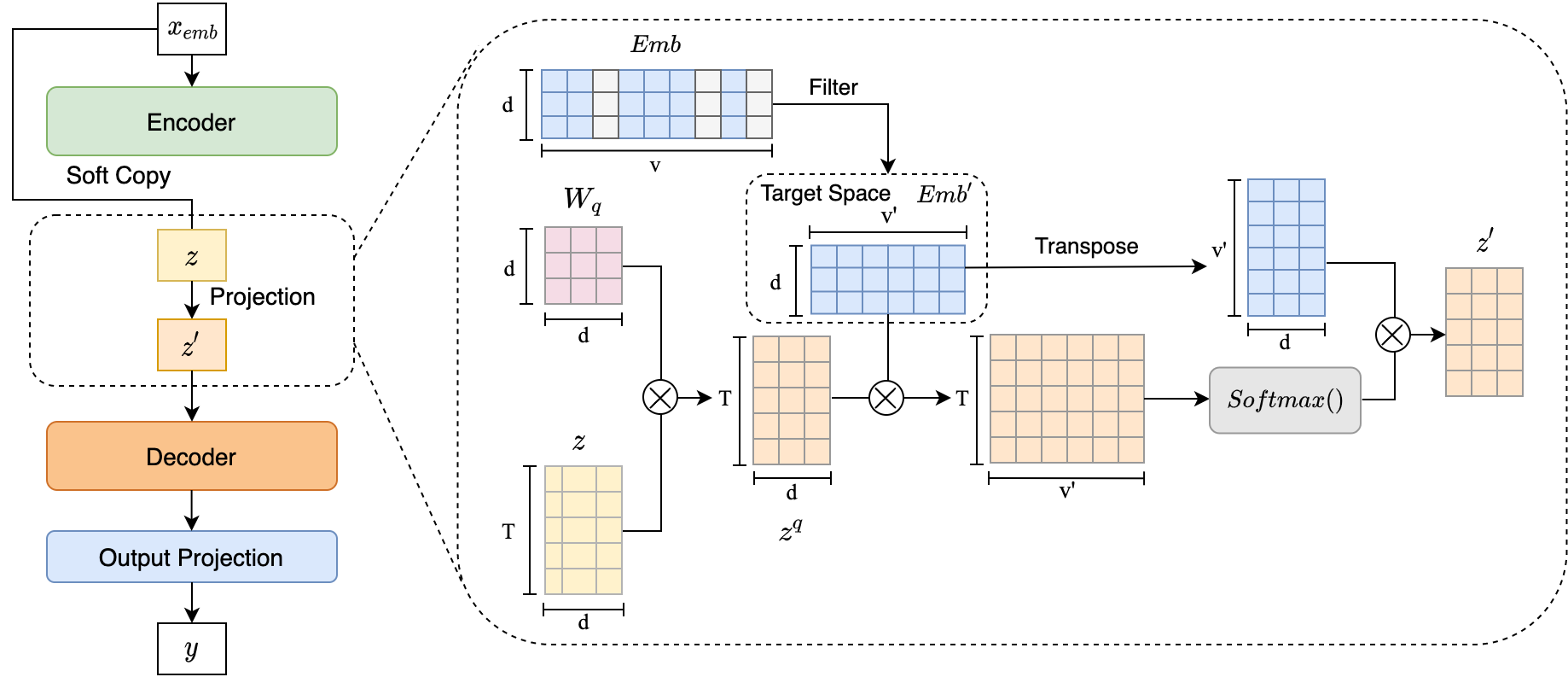}
    \caption{\small{The proposed Decoder Input Transformation \textbf{(IT)} from $z$ to $z'$, where $z\in \mathbb{R}^{T \times d}$ denotes the initial decoder input copied from the source text embedding $x_{emb}$, $T$ and $d$ denote the length of the target text $y$ and the size of hidden states, respectively. $Emb\in \mathbb{R}^{d \times v}$ denotes the output embedding matrix of the decoder (the target representation space),
    where $v$ denotes the size of the target vocabulary.}}
    \label{fig:method1}
\end{figure*}

Given the initial decoder input $z$ as a copy of source text embedding, we propose to directly select relevant representations from target embedding to form a new decoder input $z'$ (Figure~\ref{fig:method1}).   $z$ is used as the query and the selection is implemented as a learnable attention module. The learnable parameters bridge the gap between training and inference while the selection guarantees consistency between the decoder input matrix and the target representation space (i.e., the output embedding matrix of the decoder).
This way, the decoder input is in the \textit{exact} target-side embedding space and more conducive to modeling target dependencies for NAT models than previous approaches using source text embedding or transformed decoder input.

\paragraph{Decoder Input Transformation}
To transform $z$ into the target representation space, we apply attention mechanism between $z$ and the output embedding matrix $Emb \in \mathbb{R}^{d \times v} $, where $d$ and $v$ denote sizes of hidden states and the target vocabulary. Since NAT models usually have embedding matrix $Emb$ including both source and target vocabularies, first, we conduct a filtering process to remove source vocabulary (mostly not used by the decoder) from the decoder output embedding matrix (the linear layer before decoder softmax). We build a dictionary that contains only target-side tokens in the training set. We then use this dictionary to filter $Emb$ and obtain the new output embedding matrix of the decoder $Emb'\in \mathbb{R}^{d \times v'}$, where $v'$ denotes size of the filtered vocabulary. This filtering process guarantees that $Emb'$ is strictly from the target representation space.  The attention process starts with a linear transformation:
\begin{equation}
 z^{l} = W_q\cdot z
\label{eq:emb_attn}
\end{equation}
Next, the dot-product attention is performed on $z^{l}$ (as query) and $Emb'$ (as key and value): 
\begin{equation}
Sim = {\rm softmax}(z^{l}\cdot Emb')
\end{equation}
$Sim$ represents similarity between each $z^l_i$ and each embedding in the target vocabulary. 
Finally, we compute a weighted sum $z'$ of target embedding based on their similarity values:
\begin{equation}
z' = Sim\cdot Emb'^T
\end{equation}
Since $z'$ is a linear combination of $Emb'$ which is strictly in the target representation space, $z'$ is also strictly in the target representation space, hence using $z'$ as the decoder input provides a more solid basis for target dependency modeling.

\paragraph{Target-side Embedding Compression}
To reduce the computational cost of IT, we propose a target-side embedding compression approach to compress the large target embedding 
matrix. We process $Emb'$ through a linear layer to obtain a new target embedding $Emb^* \in \mathbb{R}^{d \times v^*}$:
\begin{equation}
Emb^* = (W_c \cdot Emb'^T)^T
\end{equation}
where $W_c \in \mathbb{R}^{v^* \times v'}$ is trainable and the size of compressed vocabulary $v^*$ is set manually. The result $Emb^*$
is still in the target representation space. Since we can manually set $v^*$ as a relatively small number (e.g., 1000, 2000), the computational cost of the attention mechanism can be greatly reduced. We hypothesize that target-side embedding compression may also alleviate over-fitting on small datasets and confirm this hypothesis in Section~\ref{ablation}. 

\section{Experiments}
\label{sec:experiment}
\subsection{Experimental Setup}
\label{subsec:setup}
\paragraph{Datasets}
We compare our methods with prior works on widely used MT benchmarks for evaluating NAT models: WMT14 EN$\leftrightarrow$DE (4.5M pairs), WMT16 EN$\leftrightarrow$RO (610K pairs). Also, we use IWSLT16 DE-EN (196K pairs), IWSLT14 DE-EN (153K pairs), and SP EN-JA\footnote{\url{https://github.com/odashi/small_parallel_enja}} (50K pairs) for further analysis. For WMT16 EN$\leftrightarrow$RO and IWSLT16 DE-EN, we adopt the processed data from~\cite{lee2020deterministic}. For WMT14 EN$\leftrightarrow$DE, we apply the same preprocessing and learn subwords as~\citet{gu2020fully}. For IWSLT14 DE-EN, we follow preprocessing in~\cite{guo2019non}. For SP EN-JA, we use sentencepiece\footnote{\url{https://github.com/google/sentencepiece}} to tokenize the text into subword units following \citet{chousa2019simultaneous}. Following prior works, we share the source and target vocabulary and embeddings in each language pair in $Emb$, except EN-JA. Also following prior works~\citep{gu2018non,qian2020glancing}, all NAT models in our experiments are trained on data generated from \textbf{pre-trained AT Transformer-base} with \emph{sequence-level knowledge distillation (KD)} for all datasets except EN-JA.

\paragraph{Baselines and Training}
We implement the baseline models based on their released codebases. We implement the representative vanilla NAT~\citep{gu2018non,qian2020glancing,huang2021non}\footnote{\url{https://github.com/facebookresearch/fairseq/tree/main/examples/nonautoregressive_translation}}, the highly competitive fully NAT model GLAT~\citep{qian2020glancing}\footnote{\url{https://github.com/FLC777/GLAT}}, and \textbf{current fully NAT SOTA CTC w/ DSLP \& Mixed Training} (\textbf{CTC-DSLP-MT})~\citep{huang2021non}\footnote{\url{https://github.com/chenyangh/DSLP}} and apply our methods to them. Following \citet{qian2020glancing}, we use base-Transformer ($d_{model}$=512, $n_{head}$=8, $n_{layer}$=6) for WMT datasets and small-Transformer ($d_{model}$=256, $n_{head}$=4, $n_{layer}$=5) for IWSLT and SP EN-JA datasets. We use the same training setup for training the three models, Vanilla NAT
, GLAT, and CTC-DSLP-MT as in their original papers cited above. We train models with batches of 64K tokens for WMT datasets, and 8K tokens for IWSLT and SP EN-JA datasets, using NVIDIA V100 GPUs.
For GLAT, we use Adam optimizer \citep{kingma2015adam} with $\beta$ = (0.9, 0.999) and set dropout rate to 0.1. 
For Vanilla NAT and CTC-DSLP-MT, we use Adam optimizer \citep{kingma2015adam} with $\beta$ = (0.9, 0.98).
For WMT datasets, the learning rate warms up to 5$e$-4 in 4K steps and gradually decays according to inverse square root schedule~\citep{vaswani2017attention}. As for IWSLT and SP EN-JA datasets, we adopt linear annealing (from 3$e$-4 to 1$e$-5 ) as in \citet{lee2020deterministic}. 
We choose the model with the best performance on the validation set as the final model and evaluate the final model on the test sets. For experiments using our method FBD (Section~\ref{sec:fbd}), we use the \textbf{FBF-NAT configuration} (as in Section~\ref{ablation}) and train the same number of steps at each phase (including NAT training phase), with 300K steps for each phase for WMT datasets and 100K steps for each phase for IWSLT datasets and SP EN-JA.
IT by default is \textbf{without Target-side Embedding Compression} (Section~\ref{sec:IT}).

\paragraph{Evaluation}
To evaluate the translation accuracy, we use \textbf{SacreBLEU}~\citep{DBLP:conf/wmt/Post18} for all experiments and ChrF~\citep{DBLP:conf/wmt/Popovic15} (also using the SacreBLEU tool) additionally for ablation study on IWSLT benchmark. To evaluate the inference latency, following \citet{gu2020fully}, we measure the wall-clock time for translating the entire WMT14 EN-DE test set with batch\_size=1 on a single NVIDIA V100 GPU, then compute the average time per sentence. We report \textbf{Speed-up} based on the inference latency of \textit{Transformer-base AT (teacher)} and fully NAT models.

\begin{table*}[th!]
\centering
\renewcommand{\arraystretch}{0.9}
\small
\resizebox{1.0\linewidth}{!}{\begin{tabular}{l|l|r|cc|cc}
\toprule
\multirow{2}{*}{\textbf{Row\#}} & \multirow{2}{*}{\textbf{Models}} & \multirow{2}{*}{\textbf{Speed-up $\uparrow$}} & \multicolumn{2}{c|}{\textbf{WMT'14}} & \multicolumn{2}{c}{\textbf{WMT'16}} \\
 & & & \textbf{EN-DE} & \textbf{DE-EN} & \textbf{EN-RO} & \textbf{RO-EN} \\
\midrule
1 & Transformer-\textit{base} (teacher) & 1.0$\times$ &  \bf{27.48}&\bf{31.39}&\bf{33.70}&\bf{34.05} \\
2& \quad + KD & 2.5$\times$ & 27.34& 30.95 &33.52 &34.01\\
\midrule
3& Vanilla NAT & 15.6$\times$ &20.36 & 24.81 & 28.47 & 29.43\\
4& \quad w/ DSLP$^*$  & 14.8$\times$& 22.72 &25.83 &30.48 & 31.46\\
5& \quad w/ DePA (\textbf{Ours})  &  15.4$\times$& \textbf{23.15} & \textbf{26.59}  & \textbf{30.78} & \textbf{31.89} \\
\midrule
6& GLAT  & 15.3$\times$ & 25.21 & 29.84 & 31.19 & 32.04 \\
7& \quad w/ DSLP$^*$  & 14.9$\times$& 25.69& 29.90 &32.36 &33.06\\
8& \quad w/ DePA (\textbf{Ours})  & 15.1$\times$& \textbf{26.43} & \textbf{30.42} & \textbf{33.07}& \textbf{33.82} \\
\midrule
10& CTC$^*$  & 15.5$\times$ & 25.72 & 29.89 & 32.89 & 33.79  \\
11& \quad w/ DSLP$^*$  & 14.8$\times$ & 26.85 & 31.16 & 33.85 & 34.24  \\
12& \quad w/ DSLP \& Mixed Training &14.8$\times$ &27.02 & 31.61 & $33.99$ & $34.42$ \\
13& \quad w/ DSLP \& Mixed Training \& w/ DePA (\textbf{Ours})  & 14.7$\times$& \bf{27.51} & \bf{31.96}& \bf{34.48} & \bf{34.77} \\
\midrule
14 & Average improvement from DSLP & - & 1.32 & 0.78 & 1.38 & 1.17 \\
15 & \textbf{Average improvement from DePA (Ours)} & - & \textbf{1.50} & \textbf{0.90} & \textbf{1.56} & \textbf{1.53} \\
\bottomrule
\end{tabular}}

\caption{\small{BLEU and Speed-up from our \textbf{DePA} and existing methods on WMT benchmark test sets. \textbf{Speed-up} is measured on WMT14 EN-DE test set. BLEUs without rescoring are reported, with the best BLEU scores in bold for each group. $\mathbf{^*}$ denotes the results are copied from previous work~\citep{huang2021non}, other results are obtained by our implementation. 
Average improvements of DSLP are re-calculated using our results, which are slightly different from Table 1 in ~\cite{huang2021non}.}}
\label{tab:main-rst}
\end{table*}

\subsection{Main Results}
Table~\ref{tab:main-rst} shows the main results on the WMT benchmarks.
For EN$\leftrightarrow$RO, we report the mean of BLEU from $3$ runs with different random seeds for Row 12-13, all with quite small standard deviations ($\le 0.16$) \footnote{WMT14 EN$\leftrightarrow$DE is much larger than WMT16 EN$\leftrightarrow$RO. Since standard deviations of BLEU from multiple runs with different random seeds on WMT14 EN$\leftrightarrow$DE are very small, $\le 0.08$~\citep{DBLP:journals/corr/abs-2205-07459}, following prior works, we report single-run BLEU on WMT14 EN$\leftrightarrow$DE to save energy.}.
We apply our proposed DePA, which includes IT and FBD, to vanilla NAT, GLAT, and the current fully NAT SOTA CTC-DSLP-MT, on WMT, IWSLT, and EN-JA benchmarks. We use the same hyperparameters and random seeds 
to fairly compare two models.
\textbf{It is crucial to point out that accuracies of vanilla NAT, GLAT, and CTC-DSLP-MT models have plateaued out after 300K training steps on WMT datasets hence original papers of these three models set max training steps to 300K}. We verify this observation in our own experiments as we also see no gains on these models after 300K training steps on the WMT datasets. Hence, although our DePA trains $300K\times4=1200K$ steps on WMT datasets due to \textbf{FBF} pre-training as in Section~\ref{ablation}, \textbf{all comparisons between baselines w/ DePA and w/o DePA are fair comparisons}. Table~\ref{tab:main-rst} shows that DePA consistently improves the translation accuracy for both vanilla NAT and GLAT on each benchmark, achieving  \textbf{mean=+1.37 and max=+1.88} BLEU gain on GLAT and \textbf{mean=+2.34 and max=+2.46} BLEU gain on vanilla NAT. DePA also improves the SOTA CTC-DSLP-MT by \textbf{mean=+0.42 and max=+0.49} BLEU gain on the WMT test sets (Table~\ref{tab:main-rst}), \textbf{+0.85} BLEU gain on the IWSLT16 DE-EN validation set and \textbf{+1.43} BLEU gain on the EN-JA test set (Table~\ref{tb:ablation}). 
All gains from DePA on vanilla NAT, GLAT, and CTC-DSLP-MT are \textbf{statistically significant} ($p < 0.05$) based on a paired bootstrap resampling test conducted using 1K resampling trials and the SacreBLEU tool.

Table~\ref{tab:main-rst} also shows that on each benchmark, the average improvement from DePA on three models (vanilla NAT, GLAT, and CTC-DSLP-MT) is within \textbf{[0.90,1.56]} (Row15), always larger than the average improvement from w/DSLP on them,  \textbf{[0.78,1.38]} (Row14). DePA brings consistent improvement over SOTA CTC-DSLP-MT on all benchmarks (Table~\ref{tab:main-rst} Row13-over-Row12, Table~\ref{tb:ablation}), hence we expect DePA to also improve DA-Transformer~\citep{DBLP:journals/corr/abs-2205-07459} and DDRS w/ NMLA~\citep{shao2022non} and will verify this w/ and w/o KD in future work. \textbf{Applying DePA to fully NAT models retains the inference speed-up advantages of fully NAT models.} Applying DePA to vanilla NAT, GLAT, and SOTA CTC-DSLP-MT obtain $\mathbf{{15.4}\times}$, $\mathbf{{15.1}\times}$, and $\mathbf{{14.7}\times}$ speed-up over the autoregressive Transformer-base (teacher) (Row1). Overall Table~\ref{tab:main-rst} shows that \textbf{DePA achieves greater BLEU gains with less speed-up loss than DSLP on all baselines.} These results demonstrate superiority of DePA over DSLP on improving other fully NAT models.


\begin{table*}[!tbp]
\centering
\renewcommand{\arraystretch}{0.9}
\small
\resizebox{0.85\linewidth}{!}{\begin{tabular}{l | cc |cc |cc}
\toprule
\multirow{3}{*}{\textbf{Models}} & \multicolumn{2}{c|}{\textbf{IWSLT16}}
& \multicolumn{2}{c|}{\textbf{WMT'14}} & \multicolumn{2}{c}{\textbf{WMT'16}} \\
& \multicolumn{2}{c|}{\textbf{DE-EN}} & \textbf{EN-DE} & \textbf{DE-EN} & \textbf{EN-RO} & \textbf{RO-EN} \\ 
& \textbf{BLEU} & \textbf{ChrF} & \multicolumn{2}{c|}{\textbf{BLEU}} & \multicolumn{2}{c}{\textbf{BLEU}} \\
\midrule
CTC-DSLP-MT & 31.04 & 56.7 & 27.02 & 31.61 & 34.17 & 34.60 \\
CTC-DSLP-MT w/ IT & 31.29 & 57.1 & 27.21 & 31.78 & 34.32 & 34.71 \\
CTC-DSLP-MT w/ FBD & 31.73 & 57.5 & 27.44 & 31.90 & 34.60 & 34.92 \\
CTC-DSLP-MT w/ IT+FBD & \textbf{31.89} & \textbf{57.8} & \textbf{27.51} & \textbf{31.96} & \textbf{34.68} & \textbf{34.98} \\
\midrule
\multirow{3}{*}{\textbf{Models}} & \multicolumn{2}{c|}{\bf{IWSLT16}} & \multicolumn{4}{c}{\bf{EN-JA}} \\
& \multicolumn{2}{c|}{\bf{DE-EN}} & \multicolumn{4}{c}{\textbf{}} \\
& \textbf{BLEU} & \textbf{ChrF} & \multicolumn{4}{c}{\bf{BLEU}} \\
\midrule
GLAT &  29.61 &  51.8 & \multicolumn{4}{c}{27.67} \\
GLAT (400K step) & 29.68 &  52.1 & \multicolumn{4}{c}{--} \\
GLAT w/ IT &  29.95 & 52.8 & \multicolumn{4}{c}{27.95} \\
GLAT w/ FBD & 30.87& 53.3 & \multicolumn{4}{c}{28.87} \\
GLAT w/ IT+FBD & \textbf{31.01} & \textbf{54.5} & \multicolumn{4}{c}{\textbf{29.10}} \\
\bottomrule
\end{tabular}}
\caption{\small{Effect of \textbf{IT} and \textbf{FBD} and IT+FBD (i.e., \textbf{DePA}) on the IWSLT16 DE-EN validation set, the WMT and SP EN-JA test sets. We report \textbf{mean of BLEU/ChrF} from $3$ runs with different random seeds. BLEU gains from DePA on SOTA CTC-DSLP-MT on each set,  \textbf{[0.85, 0.49, 0.51]}, are larger than std ($\le 0.17$).}}
\label{tb:ablation}
\end{table*}

\subsection{Analysis}
\label{ablation}

\begin{table}[!tbp]
\centering
\small
\resizebox{1.0\linewidth}{!}{\begin{tabular}{lc}
\toprule
\bf{Models} & \bf{BLEU} \\
\midrule
Vanilla NAT~\citep{guo2019non} & 22.95 \\
Vanilla NAT w/ Linear Mapping~\citep{guo2019non} & 24.13 \\ \hline
Vanilla NAT (our implementation) & 23.26 \\
Vanilla NAT w/ IT & \bf{26.44}  \\
\bottomrule
\end{tabular}}
\caption{\small{Compare \textbf{IT} and Linear Mapping~\citep{guo2019non} on vanilla NT on the IWSLT14 DE-EN test set.}}
\label{tb:ablation2}
\end{table}

\paragraph{Ablation Study}
We analyze the respective efficacy of IT and FBD in DePA on the IWSLT16 DE-EN validation and the WMT and SP EN-JA test sets.
Table~\ref{tb:ablation} shows that FBD and IT improve GLAT by \textbf{+1.26 BLEU/+1.5 ChrF} and \textbf{+0.34 BLEU/+1.0 ChrF} on IWSLT16 DE-EN validation set, respectively. Considering that GLAT w/FBD has more training steps than GLAT, we also train GLAT (400K steps) which has the same training steps as GLAT w/FBD for fair comparison. \textbf{Similar to findings on WMT datasets, we observe plateaus of accuracy on IWSLT and EN-JA datasets from more training steps than the original 100K}.
Just training more steps hardly improves the baseline (only +0.07 BLEU gain) on IWSLT16 DE-EN, whereas GLAT w/FBD brings \textbf{+1.19 BLEU/+1.2 ChrF} gains over GLAT (400K steps). 

Table~\ref{tb:ablation2} shows our IT outperforms Linear Mapping~\citep{guo2019non} by \textbf{+2.31} BLEU gain on IWSLT14 DE-EN test set. IT has the same number of extra parameters as Linear Mapping. Hence, the large gain proves that improvements from IT are not just from additional layers. The number of extra parameters of IT, as from $W_q$ in Eq.\ref{eq:emb_attn}, is quite small: 512*512=262144 for Transformer-base on WMT datasets and 256*256=65536 for Transformer-small on IWSLT datasets. The large BLEU gain \textbf{+3.18} from applying IT to vanilla NAT proves vanilla transformer decoder cannot achieve similar transformation effectiveness as IT. 
Table~\ref{tb:ablation} shows that for language pairs with different levels of source-target vocabulary sharing, such as WMT EN-DE and DE-EN, IWSLT DE-EN, EN-RO, and EN-JA, our IT method can achieve consistent improvements over GLAT and CTC-DSLP-MT. Applying IT consistently improves GLAT and CTC-DSLP-MT although these gains are smaller than gain on vanilla NAT. This is because decoder input of vanilla NAT only replicates source embedding,  whereas GLAT and CTC-DSLP-MT already transform decoder input by replacing selected positions in decoder input with target embedding,  hence reducing improvements of IT. Still, gains from w/IT+FBD over w/FBD confirms our hypothesis that IT can enhance effectiveness of FBD.  On GLAT, IT+FBD yields \textbf{+1.4 BLEU/+2.7 ChrF} gains on IWSLT16 DE-EN and \textbf{+1.43 BLEU} on EN-JA and on SOTA CTC-DSLP-MT,  \textbf{+0.85 BLEU/+1.1 ChrF} gain on IWSLT16 DE-EN. 

To further analyze IT, we compare cosine similarity between the target embedding against the original decoder input and the transformed decoder input, respectively.
For each sample in the IWSLT16 DE-EN validation set, we average all its token embeddings as the decoder input representation and the same for the target representation and then compute cosine similarity. We average similarities of all samples as the final similarity. 
We find that \textbf{IT significantly improves similarity between the decoder input and the target representation}, $0.04951\rightarrow0.14521$ for GLAT and $0.04837\rightarrow0.14314$ for vanilla NAT.

\paragraph{Impact of Different Dependency Curricula in FBD}
Table~\ref{tab:combine} presents results from applying different forward-backward dependency modeling curricula (Figure~\ref{fig:fbd}) on GLAT on the IWSLT16 DE-EN validation and the SP EN-JA test sets. 
Compared with modeling backward dependency in Phase 1 (B-NAT and BF-NAT), modeling forward dependency in Phase 1 (F-NAT , FB-NAT, and FBF-NAT) performs notably better. FB-NAT outperforms BF-NAT by \textbf{+3.04} BLEU on IWSLT16 DE-EN and \textbf{+2.08} BLEU on EN-JA. It seems that forward dependency modeling achieves good initialization for subsequent training phases, while backward dependency modeling cannot. 
We observe the best curriculum as \textbf{FBF-NAT}, i.e., first learn forward dependency, next learn backward dependency, then another round of forward dependency training before NAT training.
Table~\ref{tab:combine} shows the same trend of curricula on SP EN-JA as on IWSLT16 DE-EN, with FBF-NAT performing best, demonstrating that this trend of forward-backward dependency modeling curricula is consistent for both right-branching (English) and left-branching  (Japanese) target languages. All these observations confirm our hypotheses in Section~\ref{sec:fbd}. Our FBF-NAT consistently outperforms baseline GLAT (denoted by NAT in Table~\ref{tab:combine}) by \textbf{+1.58} on IWSLT16 DE-EN and \textbf{+1.56} on SP EN-JA and outperforms prior works modeling forward dependency only~\citep{guo2020fine} on GLAT (denoted by F-NAT in Table~\ref{tab:combine}) by \textbf{+1.15} on DE-EN and \textbf{+1.08} on EN-JA.

\paragraph{DePA on Raw Data}
We evaluate DePA on raw data by training models on the original training set without KD (Section~\ref{subsec:setup}). DePA improves GLAT on the IWSLT16 DE-EN validation set by \textbf{+1.57} BLEU (26.57 $\rightarrow$ 28.14),  proving that DePA effectively enhances the dependency modeling ability of fully NAT models hence \textbf{reduces dependence of NAT training on AT models}.

\begin{table*}[t]
\small
\centering
\renewcommand{\arraystretch}{0.9}
\resizebox{0.85\linewidth}{!}{
\begin{tabular}{ lc | lc |  lc | lc }
\toprule
\multicolumn{4}{c|}{\bf{IWSLT16 DE-EN validation set}} & \multicolumn{4}{|c}{\bf{SP EN-JA test set}} \\
\midrule
\bf{Models} & \bf{BLEU} & \bf{Models} & \bf{BLEU} & \bf{Models} & \bf{BLEU} & \bf{Models} & \bf{BLEU} \\
\midrule
NAT & 29.61& BF-NAT & 27.83  & NAT & 27.67 & BF-NAT & 26.79 \\
F-NAT & 30.04& FB-NAT & 30.87 & F-NAT & 28.15& FB-NAT & 28.87\\
B-NAT & 27.05& FBF-NAT& \bf{31.19} & B-NAT & 25.83 & FBF-NAT& \bf{29.23}\\
\bottomrule
\end{tabular}}
\caption{\small{BLEU from different dependency modeling curricula on GLAT. Best results for each set are in bold. \textbf{NAT} denotes GLAT baseline. \textbf{F} and \textbf{B} denote forward dependency and backward dependency phase respectively (Figure~\ref{fig:fbd}). For example, F-NAT denotes forward dependency training then NAT training.}}
\label{tab:combine} 
\end{table*}

\begin{figure}[hbt]
    \centering
    \begin{subfigure}[b]{0.225\textwidth}
        \centering
        \includegraphics[width=\textwidth]{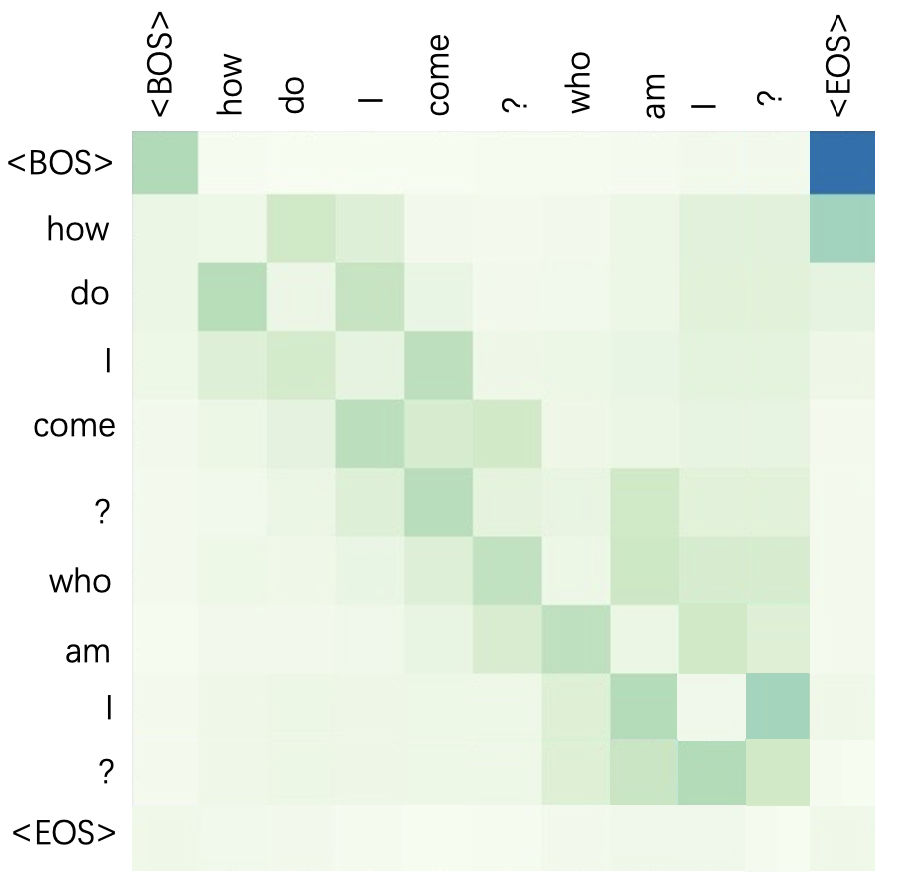}
        \caption{NAT}
        \label{fig:nat}
    \end{subfigure}
    \hfill
    \begin{subfigure}[b]{0.225\textwidth}
        \centering
        \includegraphics[width=\textwidth]{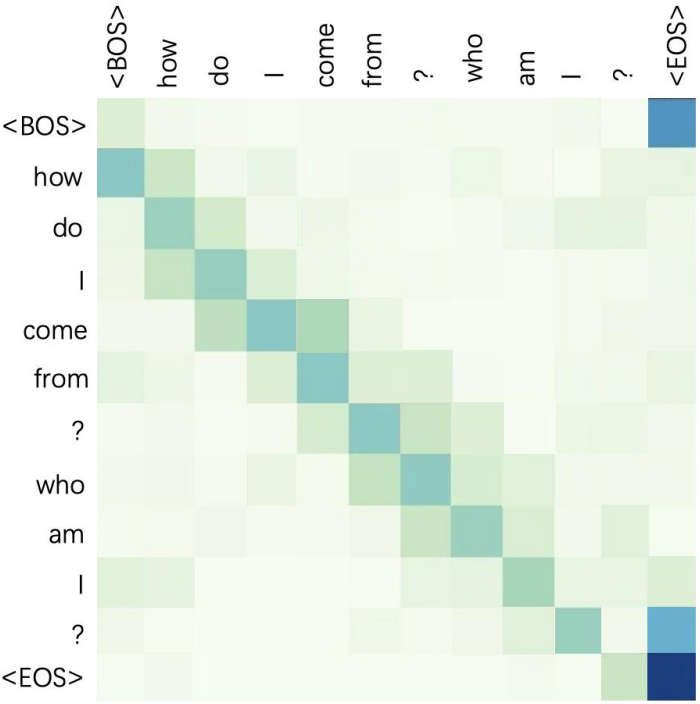}
        \caption{F-NAT}
        \label{fig:f-nat}
     \end{subfigure}
    \hfill
     \begin{subfigure}[b]{0.225\textwidth}
        \centering
        \includegraphics[width=\textwidth]{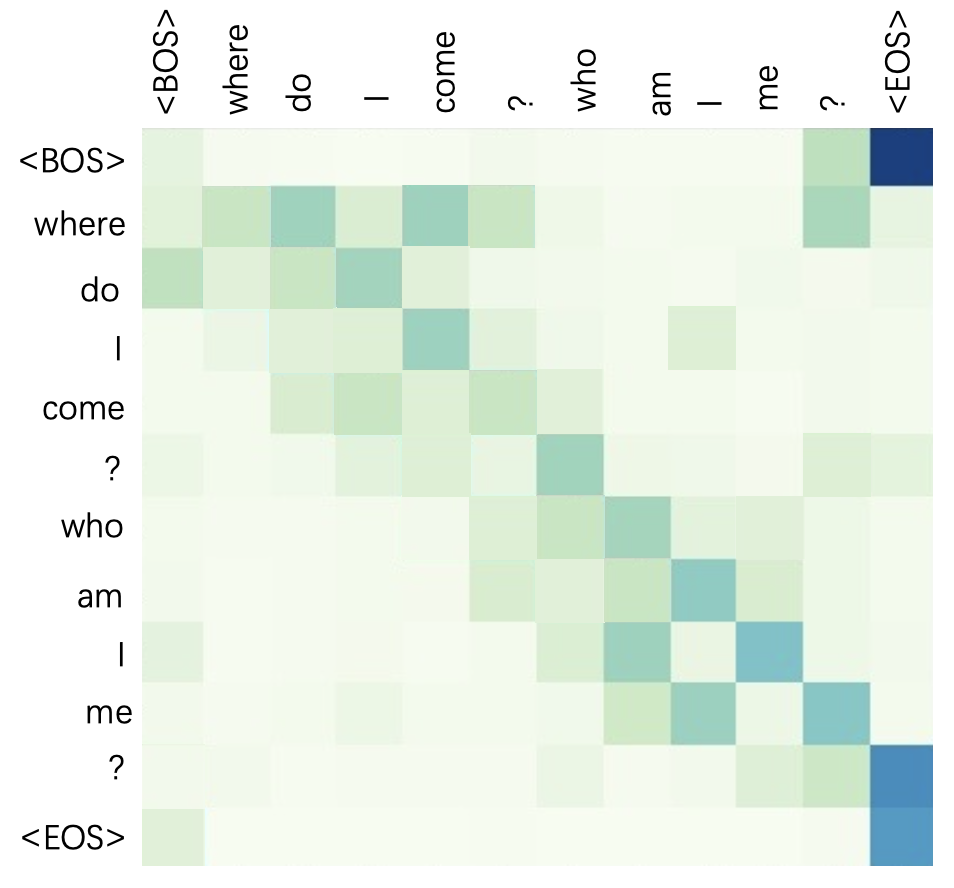}
        \caption{B-NAT}
        \label{fig:b-nat}
     \end{subfigure}
     \hfill
    \begin{subfigure}[b]{0.225\textwidth}
         \centering
         \includegraphics[width=\textwidth]{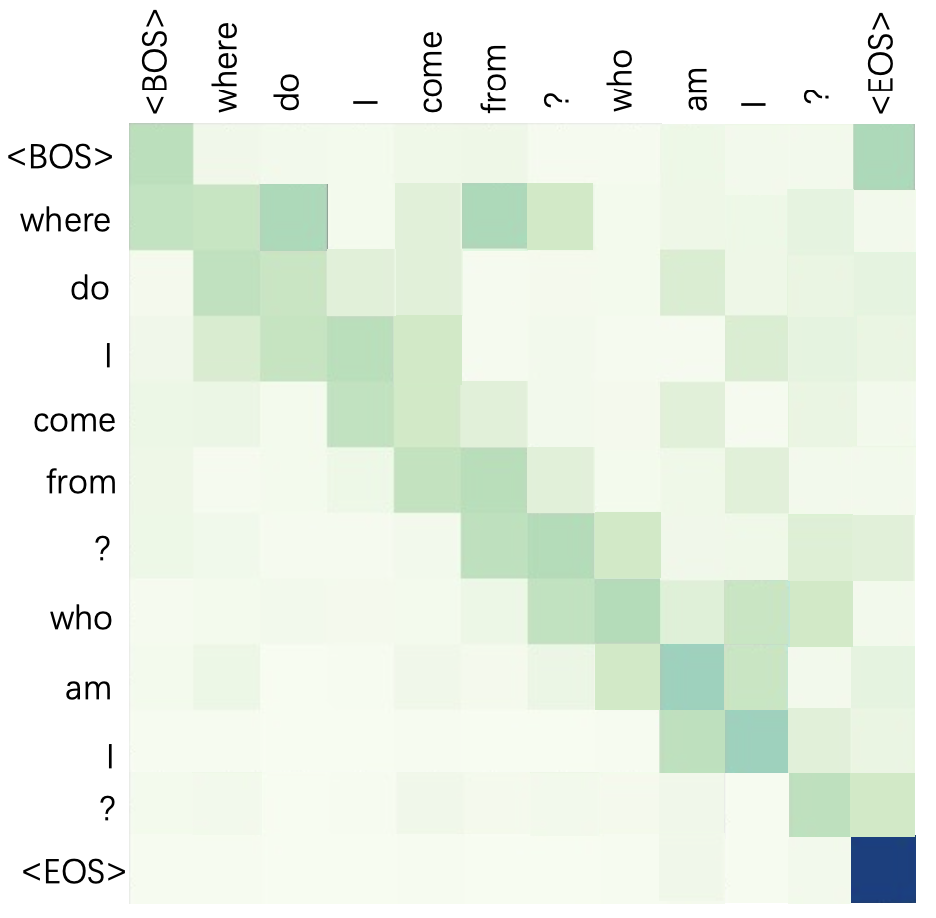}
         \caption{FB-NAT}
         \label{fig:fb-nat}
     \end{subfigure}
        \caption{\small{Visualization of the decoder self-attention distribution in NAT models on IWSLT16 DE-EN validation set. Definitions of model names are the same as in Table~\ref{tab:combine}.}}
        \label{fig:attn_map}
\end{figure}

\paragraph{Effectiveness of Target-side Embedding Compression} 
We propose a linear compression module to reduce the selection candidates of the target embedding for IT (Section~\ref{sec:IT}). We use the dichotomy to determine the compression dimension interval $[1000 , 2000]$ and evaluate GLAT w/ IT using different dimensions with step size 200 in this interval for IT on the IWSLT16 DE-EN validation set.
As shown in Table~\ref{tab:compress},
applying IT on GLAT improves BLEU up to \textbf{+0.78}    (29.61$\rightarrow$ 30.39) with compressed dimension 1800. We also experiment with target-side embedding compression on a larger model on WMT16 EN-RO but find no gains. We assume that for relatively small models and data, this approach helps filter out some redundant target information, hence refines the target representation space and improves the translation accuracy.
\begin{table}[hbpt]
\small
\centering
\resizebox{1.0\linewidth}{!}{
\begin{tabular}{c|rrrrrrr}
\toprule
\textbf{Compressed}& \multirow{2}*{w/o IT} &\multirow{2}*{1000}&\multirow{2}*{1200}&\multirow{2}*{1400}&\multirow{2}*{1600}&\multirow{2}*{1800}&\multirow{2}*{2000} \\
\textbf{Dimension}&&&&&&&\\
\midrule
\textbf{BLEU}& 29.61&29.45&29.56&29.77&29.85&\bf{30.39}&29.14\\
\bottomrule
\end{tabular}}
\caption{BLEU from GLAT w/ IT on the IWSLT16 DE-EN validation set with Target-side Embedding Compression described in Section~\ref{sec:IT}.}
\label{tab:compress} 
\end{table}

\subsection{Case Study and Visualization}
\label{appendix:visualization}
Table~\ref{tab:case_study_fbd} presents case studies of GLAT w/ FBD (``FB-NAT'') and with only forward modeling (``F-NAT'') on IWLST16 DE-EN validation set. Some typical multi-modality errors in F-NAT predictions are corrected by incorporating both forward and backward dependency modeling through FBD. 

For a more intuitive analysis of FBD, we present a visualization of the decoder self-attention distribution of different NAT models in Figure~\ref{fig:attn_map}. All models are based on GLAT and model names conform to those in Table~\ref{tab:combine}. In the baseline GLAT (Figure~\ref{fig:nat}), the self-attention distribution of each position is scattered in adjacent positions, indicating that the NAT model lacks dependency and has high confusion during decoding, causing multi-modality errors. In F-NAT and B-NAT models, significant forward and backward dependencies can be observed in Figure~\ref{fig:f-nat} and \ref{fig:b-nat}, indicating that these two models can better use information in previous or future positions. Encouragingly, forward and backward dependencies are fused in the FB-NAT model (Figure~\ref{fig:fb-nat}), which can focus on future information while modeling forward dependency, capable of alleviating problems shown in Table~\ref{tab:case_study_fbd}. 

\section{Conclusion}

We propose a novel and general Dependency-Aware Decoder (DePA) to enhance target dependency modeling for fully NAT models, with forward-backward dependency modeling and decoder input transformation.
Extensive experiments show that DePA improves the translation accuracy of highly competitive and SOTA fully NAT models while preserving their inference latency. In future work, we will evaluate DePA on iterative NAT models such as Imputer, CMLM, and Levenshtein Transformer and incorporate ranking approaches into DePA.
\newpage

\section{Limitations}
Apart from all the advantages that our work achieves, some limitations still exist.
Firstly, in this work, we investigate the efficacy of applying our proposed DePA approach on the representative vanilla NAT, the highly competitive fully NAT model GLAT and current SOTA CTC-DSLP-MT for fully NAT models, but we have yet to apply DePA to iterative NAT models, such as Imputer~\cite{saharia2020non}, CMLM~\cite{ghazvininejad2019mask}, and Levenshtein Transformer~\cite{gu2019levenshtein}.
Hence, the effectiveness of DePA on iterative NAT models still needs to be verified.  Secondly, we have not yet incorporated re-ranking approaches such as Noisy Parallel Decoding (NPD)~\cite{gu2018non} into DePA.
Thirdly, our proposed method FBD requires multiple additional training phases before NAT training, resulting in longer training time and using more GPU resources. Reducing the computational cost of FBD training is one future work that will be beneficial for energy saving.  Last but not least, NAT models have limitations on handling long text. They suffer from worse translation quality when translating relatively long text. 
 We plan to investigate all these topics in future work.

\bibliography{anthology}

\begin{table*}[htb]
\small
\centering
\resizebox{1.0\linewidth}{!}{
\begin{tabular}{c|l|l}
\toprule
& \multicolumn{1}{c|}{Case \#1} & \multicolumn{1}{c}{Case \#2} \\
\hline
\multirow{2}{*}{Source} & obwohl sie erwischt wurden , wurden sie \underline{\textbf{schließlich}} & \multirow{2}{*}{das ist ein Bauplan für Länder wie China und den Iran .} \\
& freigelassen aufgrund immensen internationalen Drucks . & \\
\hline

\multirow{2}{*}{Target Reference} &  even though they were caught , they were eventually  &\multirow{2}{*}{this is a blueprint for countries like China and Iran .} \\
& released after heavy international pressure . & \\
\hline

\multirow{2}{*}{Vanilla NAT} & although they were caught , they were {\color{red}released released} & \multirow{2}{*}{this is a blueprint {\color{blue}plan} for countries like China {\color{red}and and} Iran .} \\
& {\color{red}because because} of huge {\color{blue}drug} . & \\
\hline
\multirow{2}{*}{\textbf{Vanilla NAT w/ IT}} &  although they were caught , they were finally released & \multirow{2}{*}{this is a blueprint for countries like China and Iran .} \\
& because huge international pressure . & \\
\hline
\multirow{2}{*}{GLAT} & although they were caught , they finally were released & \multirow{2}{*}{this is a blueprint {\color{blue}plan} for countries like China and Iran .} \\
& because {\color{red}of of} international  {\color{blue}printing} . & \\
\hline

\multirow{2}{*}{\textbf{GLAT w/ IT}} & although they were caught , they were finally & \multirow{2}{*}{this is a blueprint for countries like China and Iran .} \\
& released after huge international pressure . &\\
\bottomrule
\end{tabular}}
\caption{Case studies of our method \textbf{IT} on the IWSLT16 DE-EN validation set by comparing the translations from the two baseline models Vanilla NAT and GLAT and from them after applying IT (models in bold). Repetitive tokens are in red. Source words that are not semantically translated are marked in bold and underlined (under-translation). Wrong lexical choice (incorrect translations caused by polysemy) and redundant words are in blue.}
\label{tab:case_study}
\end{table*}

\paragraph{Case study for the proposed IT}
NAT models generally suffer from the multi-modality problem, which shows as \emph{over-translation} (repetition), \emph{under-translation} (missing information), and \emph{wrong lexical choice} (incorrect translations caused by polysemy)~\cite{ran2020learning}. As shown in Table~\ref{tab:case_study}, Vanilla NAT and GLAT tend to generate repetitive tokens which are highlighted in red (over-translation). Additionally, Vanilla NAT omits the translation of ``schließlich'' which is in bold and underlined (under-translation). By applying our method IT, the decoder input is closer to the target representation space and the model has a better perception for the target-side information, so that the repetition and under-translation problems can be effectively alleviated. As for incorrect translations caused by polysemy, as shown in Case \#1 in Table~\ref{tab:case_study}, ``Drucks'' means both ``printing'' and ``pressure'' in German. GLAT mistakenly translates ``Drucks'' into ``printing'', but our method can help the model correctly translate it into ``pressure''. Particularly, in Case \#2, ``Bauplan'' means ``blueprint'' in German. Although both baseline models Vanilla NAT and GLAT generate the correct words, they also generate the redundant word ``plan'' which is also a subword of ``Bauplan''. These examples demonstrate that the baseline models may confuse the source representation space with the target representation space during generation, but our method IT effectively remedies this problem.

\end{document}